\documentclass{article}

\usepackage[final, nonatbib]{neurips_2023}
\usepackage[utf8]{inputenc} 
\usepackage[T1]{fontenc}    
\usepackage{hyperref}       
\usepackage{url}            
\usepackage{booktabs}       
\usepackage{amsfonts}       
\usepackage{nicefrac}       
\usepackage{microtype}      
\usepackage{graphicx}
\usepackage{wrapfig}
\usepackage{changepage}  
\usepackage[version=4]{mhchem}
\usepackage{amsmath, amssymb}

\title{AdsorbRL: Deep Multi-Objective Reinforcement Learning for Inverse Catalysts Design}

\author{
  Romain Lacombe\\
  Stanford University
  \And
  Lucas Hendren\\
  Stanford University
  \And
  Khalid El-Awady\\
  Stanford University
  \And
  \texttt{\{rlacombe, hendren, kae\}@stanford.edu}
}

\begin{document}
\maketitle

\begin{abstract}
A central challenge of the clean energy transition is the development of catalysts for low-emissions technologies. Recent advances in Machine Learning for quantum chemistry drastically accelerate the computation of catalytic activity descriptors such as adsorption energies. Here we introduce \emph{AdsorbRL}, a Deep Reinforcement Learning agent aiming to identify potential catalysts given a multi-objective binding energy target, trained using offline learning on the \textit{Open Catalyst 2020} and \textit{Materials Project} data sets. We experiment with Deep Q-Network agents to traverse the space of all $\sim$160,000 possible unary, binary and ternary compounds of 55 chemical elements, with very sparse rewards based on adsorption energy known for only between 2,000 and 3,000 catalysts per adsorbate. To constrain the actions space, we introduce Random Edge Traversal and train a single-objective DQN agent on the known states subgraph, which we find strengthens target binding energy by an average of 4.1 eV. We extend this approach to multi-objective, goal-conditioned learning, and train a DQN agent to identify materials with the highest (respectively lowest) adsorption energies for multiple simultaneous target adsorbates. We experiment with Objective Sub-Sampling, a novel training scheme aimed at encouraging exploration in the multi-objective setup, and demonstrate simultaneous adsorption energy improvement across all target adsorbates, by an average of 0.8 eV. Overall, our results suggest strong potential for Deep Reinforcement Learning applied to the inverse catalysts design problem.
\end{abstract}

\section{Introduction: Challenges of Catalysts Design}

A central challenge of the clean energy transition is the development of high-performance catalysts for electrochemical and thermocatalytic energy conversion \cite{jaramillo2017} \cite{chu2012opportunities}. The problem of catalysts design \cite{Zitnick2020}, which seeks to identify high-performance materials with increased intrinsic activity for any desired reaction of interest, is a critical technology enabler for low-emissions technologies, from \ce{H_2} production from water and renewable energy, to the transformation of waste \ce{CO_2} into valuable non-fossil fuels and feed-stock \cite{de2019would}. Rapid progress in Machine Learning (ML) methods for computational quantum chemistry \cite{DFT2005}, along with recently available datasets of catalysts \cite{OCP} and their properties \cite{MaterialsProject}, could significantly accelerate the identification of materials with enhanced catalytic activity.

A key descriptor of catalytic activity is \textbf{adsorption energy}, or the energy with which the reagent species or reaction intermediate (\textbf{`adsorbate'}), a small molecule on a surface site (e.g. \ce{H2O$\star$}), binds to the surface of the heterogeneous \textbf{catalyst} (here an inorganic compound nano-particle). A core tenet of heterogeneous catalysis science, the Sabatier Principle \cite{Sabatier}, holds that the optimal catalyst should have a binding energy with the reactants that is neither too weak nor too strong. 

Identifying materials that best match a target energy profile for multiple adsorbates, for instance strong binding with \ce{$\star$OH} but weaker adsorption of \ce{H2O}$\star$, is thus particularly helpful for catalysts design \cite{perez-ramirez2019strategies}. Computational approaches to catalyst design have traditionally leveraged advances in computational chemistry to screen large chemical spaces for materials with optimal adsorption energies for key intermediates \cite{norskov2009computational}. Despite advances in Density Functional Theory (DFT) \cite{DFT2005} computation, estimating adsorption energy for a single (catalyst, adsorbate) pair still requires heavy computational resources, which makes \emph{in silico} high-throughput screening of catalysts costly \cite{AdsorbML}.

Inverse design adopts the opposite approach: starting from the desired property, the task aims to design catalysts from first principles so they best fit that objective \cite{freeze2019search}. While ML techniques for inverse materials design show strong promise, synthesizability and physical realization of discovered materials is a challenge \cite{noh2020machine} compared to high-throughput screening of known materials.

We turn to Reinforcement Learning (RL) \cite{reinforcement} to propose a third way: training an agent to traverse a space of materials, not by exhaustive search, but by gravitating towards optima for the target property. 

\textbf{We introduce \emph{AdsorbRL}, a Deep Reinforcement Learning (DRL) \cite{deepRL} agent trained to traverse a space of materials and identify promising catalysts given a multi-objective binding energy target}. Specifically, we use Offline RL \cite{levine2020offline} on the \textit{Materials Project} \cite{MaterialsProject} and \textit{Open Catalyst 2020} \cite{OCP} datasets of adsorption energies to train a DRL agent to identify catalysts which bind the strongest (lowest adsorption energy) or the weakest (highest adsorption energy) with an array of target adsorbates, chosen for their importance for the clean energy transition.

We hypothesize that RL can be especially helpful to navigate chemical space in the multi-objective setting. By learning to traverse a sparse rewards environment in a multi-objective goal-conditioned setup, our agent could help identify materials with desirable properties for the multi-objective target at hand, and serve as an \textit{in silico} rapid screening mechanism to identify leads on which to further focus computational chemistry resources (DFT computations, MD simulations, etc.). 

This paper presents our experiments with several Deep RL setups to traverse a large space of compounds, and identify materials with the desired adsorption energies profile for a set of adsorbates of interest. We introduce Multi-Objective DQN with Sub-Sampling, and a novel algorithm to train such a generalized multi-objective agent.

Our findings indicate the promise of Deep RL in navigating complex chemical spaces, and present novel approaches to tackle goal-conditioned multi-objective Reinforcement Learning for materials design. These methods could serve as a foundation for more complex computational challenges in large chemical spaces, and the development of novel materials for a wide range of applications in heterogeneous catalysis, electrochemical energy conversion, and low-emissions technologies.

\begin{figure}[!ht]
\centering
    \includegraphics[width=0.8\linewidth]{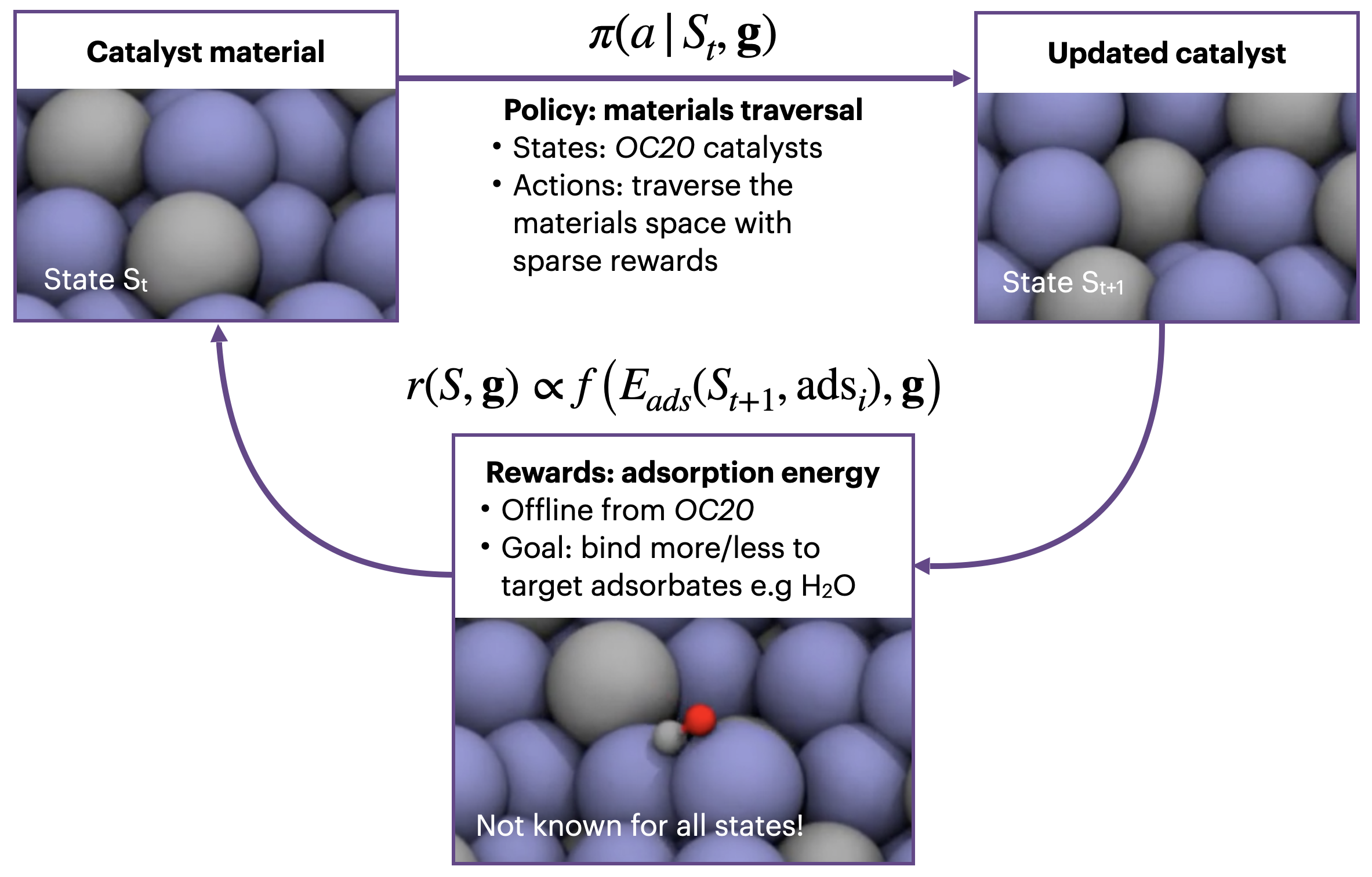}
\caption{Deep RL approach to the catalyst design problem. Images: \textit{\href{https://opencatalystproject.org/}{Open Catalyst Project}.}} 
\label{fig:RL-approach}
\end{figure}

\section{Related Work}

\subsection{Machine Learning for Adsorption Energy Predictions}

The \textit{Open Catalyst Project} \cite{OCP} is a research effort between Meta Fundamental AI Research (FAIR) and Carnegie Mellon University using Artificial Intelligence to model and discover new catalysts for renewable energy conversion and storage. Quantum mechanical computational methods such as Density Functional Theory \cite{DFT2005} can help evaluate new catalysts, but are severely limited by the very high computational cost of the higher precision levels of theory. 

For that purpose, \textit{Open Catalyst Project} has developed and released two public datasets, \emph{Open Catalyst 2020 (OC20)} \cite{OCP} and \emph{Open Catalyst 2022 (OC22)} \cite{Tran_2023}, to train ML models to efficiently approximate these calculations. These datasets together contain 1.3 million molecular relaxations with results from over 260 million DFT calculations. Importantly, Lan et al. \cite{AdsorbML} recently released \emph{AdsorbML}, an hybrid ML and computational DFT model trained to predict the adsorption energy of an adsorbate onto the sorbent surface. Inference greatly accelerates computation and achieves final energies within $\sim$0.1 eV of ground truth.

\subsection{Machine Learning for Catalysts Design}

Rapid progress in Machine Learning has lead to increased interest in applying data-driven learning techniques to the identification of catalysts. Zitnick et al. \cite{Zitnick2020} introduce the general problem of catalysts design, and provide an overview of Machine Learning approaches to the problem to motivate the introduction of the \emph{Open Catalyst} dataset. Seh et al. \cite{jaramillo2017} provide an overview of how chemists in the laboratory can combine experiment and theory to design catalysts in order to increase their intrinsic activity or their number of active sites.

Freeze et al. \cite{freeze2019search} review recent advances in inverse design for catalysts, including approaches based on Machine Learning, as well as more general optimization (gradient ascent, genetic algorithms), but do not point to Reinforcement Learning. Zhu et al. \cite{Zhu2023} and Noh et al. \cite{noh2020machine} provide a comprehensive review of the published literature on Machine Learning for electrocatalysts and inorganic solid materials, respectively, including approaches using \emph{OC20} data. While many different techniques have been applied to this problem space, none based on Reinforcement Learning are cited. 

\subsection{Reinforcement Learning for Molecular and Materials Design}

In the more general area of inverse materials design, a number of Deep RL-based approaches have reported success. Zhou, et al. \cite{Zhou2019} apply the Deep Q-Network (DQN) algorithm \cite{mnih2015human} to molecular optimization. Their work demonstrates the effectiveness of a multi-objective reinforcement learning network for organic materials design, via a step-by-step addition or removal of elements or bonds.

Sridharan, et al. \cite{Spec2Struc} explore the application of Deep RL to molecules generation by training a model to reconstruct 2D molecular graph based Nuclear Magnetic Resonance (NMR) spectroscopy.  They designed an innovative framework using Monte-Carlo-Tree-Search (MCTS) and Graph Convolutional Networks to reconstruct the most likely molecular structure based on the chemical information extracted from NMR spectroscopy output. They demonstrated that applying their model in a RL setting helped consistently identify exact molecular structures based only on spectrograms.

Pan, et al. \cite{Pan2022} apply Deep Q-Networks \cite{mnih2015human} to inverse inorganic materials design.  This paper demonstrates the use of Deep RL for the identification of new materials based on desired properties, and was especially novel in its approach of inorganic materials design modeling, and how to integrate chemical constraints through Lagrange multipliers to impose charge neutrality and electro-negativity balance. This approach demonstrated an RL agent navigating trajectories in state space via successive additions of elements to an inorganic compound for the inverse design task.

Finally Sui, et al. \cite{Sui2021} demonstrate the effectiveness of Deep RL for general digital materials design. They specifically show how RL improves on traditional approaches to find new design patterns even in vast design spaces, and show an application to additive manufacturing via the selection of soft or stiff voxels.
 
To our knowledge, Deep RL techniques have not been tried on sorbents and electrocatalyst design to date, which is what we propose in this paper.

\section{Dataset and Features}

Our dataset is sourced from the \textit{Materials Project}, an open-access database for discovery of inorganic materials, which provides pre-computed properties for a large number of candidate catalysts \cite{MaterialsProject}. Specifically, we use the \textit{Catalysis Explorer}, an online application that provides pre-computed adsorption energies for various catalysts in the \textit{Materials Project} under different configurations, sourced from the \textit{Open Catalyst Project OC20} \cite{OCP} dataset. We explore catalysts associated with the following 6 adsorbates, which are of major significance for clean energy: 
\begin{itemize}
    \item \textbf{\ce{$\star$OH2}}: adsorbed water, a key reagent for the Oxygen Evolution Reaction (OER) in \ce{H2} generation from water electrolysis \cite{man2011universality}, and a key product of the Oxygen Reduction Reaction (ORR) reaction for \ce{H2} fuel cells \cite{shao2016recent};
    \item \textbf{\ce{$\star$OH}}: adsorbed hydroxyl radical, a key intermediate for OER \cite{man2011universality} and ORR \cite{shao2016recent}, often a rate-limiting step requiring the application of electrochemical over-potentials as its adsorption energy tends to scale linearly with those of \ce{$\star$O} and \ce{$\star$OOH} \cite{perez-ramirez2019strategies};
    \item \textbf{\ce{$\star$CH4}}: adsorbed methane, of importance for direct air capture of natural gas, and for the \ce{CO2} Reduction Reaction (\ce{CO2}RR) for e.g. methanation of captured carbon dioxide \cite{nitopi2019progress}; 
    \item \textbf{\ce{$\star$CH2}}: adsorbed methylene radical, another important intermediate for \ce{CO2} Reduction Reaction (\ce{CO2}RR) for the ethylene electro-catalystic  production pathway, a key building block of the modern chemicals industry \cite{de2019would};
    \item \textbf{\ce{$\star$N2}}: adsorbed molecular nitrogen, a key reagent for the Nitrogen Reduction Reaction (NRR) \cite{singh2017electrochemical} for ammonia production, an essential ingredient of synthetic fertilizers without which half the world population would not be fed \cite{erisman2008century};
    \item \textbf{\ce{$\star$NH3}}: adsorbed ammonia, the desired product of Nitrogen reduction (NRR) \cite{singh2017electrochemical}.

\end{itemize}
    
For each adsorbate compound, the \textit{Materials Project Catalysis Explorer} provides a set of properties including the compound, bulk formula, adsorption energy, and Miller indices describing a particular lattice structure and orientation. For tractability purposes, and reasoning that the lowest energy configuration may be most representative of overall activity, we filter for all unique 1, 2, and 3-element compounds and select the \textbf{lowest adsorption energy configuration} for the target adsorbate among all combinations of stoichiometry and Miller indices, ignoring lattice orientation, and coordination environment.

This simplifies the materials space exploration problem to \textbf{traversal of the space of all $\sim$160,000 possible unary, binary and ternary compounds of 55 chemical elements} (Figure \ref{fig:periodicTableExp}). Adsorption energies are known for only between 2,000 and 3,000 catalysts per adsorbate, providing a Given the \textbf{very sparse reward} for our agent, with only 7,386 total unique catalysts in our dataset of $\sim$160,000 possible compounds with known adsorption energy for at least one adsorbate (Table \ref{tab:dataset}). We use the $\star\ce{OH2}$ adsorbate for our single objective experiments, and the 6 target adsorbates above for the multi-objective setup. Finally, in experiments (3), (4) and (5), we \textbf{further constrain this problem to the \emph{OC20}-subgraph} of compounds with known adsorption energy for at least one target adsorbate. 

Our dataset is available for download on the \href{https://github.com/rlacombe/AdsorbRL/tree/main/data}{AdsorbRL GitHub repository}.

\section{Model: Reinforcement Learning Setting} 
\label{sec:RL}

Our Reinforcement Learning model comprises of the following elements, illustrated in Figure \ref{fig:RL-approach}.

\textbf{States: $S$}. States are unary, binary or ternary compounds of 55 elements matching to catalysts from the \textit{Materials Project} dataset, comprising of 1 to 3 atomic elements forming the compound (e.g. $\ce{SiC}$: Silicon carbide). Catalysts are represented by a 55 dimensional one-hot vector in our full generality setup (experiments (1), (3), (4), and (5)). In experiment (2), states are simply a one-element square on the Periodic Table of Elements grid, represented by their atomic number $Z$.

\textbf{Actions: $a$}. These are steps the agent can take to traverse the dataset of materials. They are defined differently for each experimental setup:
\begin{itemize}
    \item Experiment (1): in the full states/full actions setup (section \ref{sec:exp1}), actions can be: addition of an element (subject to the total elements in the catalyst being less than or equal to 3); removal of an element (subject to there being at least one element in the catalyst); `do-nothing',  an action that leaves the catalyst as is; and a terminate action that ends the episode. 
    \item Experiment (2): on the Periodic Table setup (section \ref{sec:exp2}), only 5 actions are allowed: move left by one element, move right by one element, move up by one element, move down by one element, or stay in place. Trajectories automatically terminate after 9 steps.
    \item Experiment (3)---(5): in the full states/constrained actions setup (section \ref{secRandomEdges} and \ref{secMulti}), there are five possible actions: add a random element (`random edge'), remove the first, second, or third element in the catalyst, or terminate the episode.
\end{itemize}

\textbf{Reward: $r \propto f \left(E_{ads}(S_i)\right)$.} The reward for arriving in a state is a function of its adsorption energy ($E_{ads}$) for the target adsorbate. In single-objective experiments, we use $r = - E_{ads}$ and $r = E_{ads}^2$ to target terminal states with the strongest binding possible (measured by a large, negative energy value). For multi-objective setup we use $r = - E_{ads}^3$ to encourage the agent to find extrema with strongest (respectively, weakest) adsorption. Rewards functions for each experiment are reported in Table \ref{tab:experiments}.

\textbf{Termination:} we either offer termination as an action option to the agent, or upon completion of a number of steps (e.g 9 steps). 

\textbf{Policy: $\pi(a|S) $.} We train a policy to choose the next action at each state, using the Q-Learning \cite{sutton2018reinforcement} and Deep Q-Network algorithms \cite{mnih2015human} with the following Bellman equation \cite{bellman1954theory}:
$$ \mathrm{Q}^*(a|S) = r(a|S) + \gamma \max_{a} \left( \mathrm{Q}^*(a|S') \right) $$
\textbf{Multi-objective goal-conditioning.} In this setup, we train our policy to follow multiple objectives at once. We define an objective vector $\mathbf{g}$ as an array of +1 or -1 for each of the 6 adsorbates in our dataset, encoding whether we seek to bind stronger (+1, minimize energy) or weaker (-1, maximize energy). The reward becomes a function of $E_{ads}$ for each adsorbate and the objective vector $\mathbf{g}$:
$$ \pi(a|S) = \pi (a| S, \mathbf{g})$$
$$\mathbf{g} = (g_i) = (\dots, \{ -1, +1 \}_i, \dots) $$
$$r(S, \mathbf{g}) = f\left( E_{ads}(S_{t+1}, \mathrm{ads}_i) \right), \mathbf{g}) $$

The Bellman equation \cite{bellman1954theory} in the goal-conditioned, multi-objective setup becomes:
$$ \mathrm{Q}^*(a|S, \mathbf{g}) = r(a|S,\mathbf{g}) + \gamma \max_{a} \left( \mathrm{Q}^*(a|S', \mathbf{g}) \right) $$

\textbf{Evaluation metrics.} To evaluate our trained policies, we run a number of roll-outs from random initial states and compute the average adsorption energy of all final states reached by the policy. The delta with the average energy of initial states (respectively its inverse for weak binding targets in the multi-objective setup) measures how well our agent solves the problem:
$$ \Delta = \frac{1}{N} \sum_{i \in \textrm{final}} - \left( E_{ads}(S_i) \right) - \frac{1}{N} \sum_{j \in \textrm{initial}} - \left(E_{ads}(S_j) \right)$$

\section{Experiments \& Results}

We present the setup for each of the five experiments we report in Table \ref{tab:experiments}.

\begin{table}[!ht]
\centering
\resizebox{\textwidth}{!}{%
\begin{tabular}{lllllc}
Exp.  & Target adsorbates & States & Actions & Algorithm & Reward \\ \hline \\
(1)  & Single ($\star$\ce{OH2}) & All states & Graph traversal & DQN & $-
E_{ads}$ \\
(2) & Single ($\star$\ce{OH2}) & Periodic table & $\leftarrow | \rightarrow | \uparrow | \downarrow$ & Q-Learning & $E_{ads}^2$ \\ 
(3)  & Single ($\star$\ce{OH2}) & \emph{OC20}-subgraph & Random edges & DQN & $-
E_{ads}^3$ \\ 
(4)  & Multi-objective & \emph{OC20}-subgraph & Random edges & DQN & $-
E_{ads} \cdot \mathbf{g}$ \\ 
(5)  & Multi-objective & \emph{OC20}-subgraph & Random edges & DQN with Sub-Sampling & $-
E_{ads_i} g_i \quad i \sim \mathcal{U}(i=1\dots6)$ \\ \\ \hline  \\
\end{tabular}%
}

\label{tab:experiments}
\caption{Summary of experiments (1) to (5).}
\end{table}

\subsection{Offline RL on Full State \& Actions Space}
\label{sec:exp1}
\paragraph{Experiment (1).}
Our first experiment attempts to train an agent to traverse the full compounds space from a random initial catalyst to the lowest energy catalyst using reward shaping \cite{Wu2019}. Our offline data represents the set of all possible tuples, $(s, a, r, s')$ where $s$ and $s'$ are existing catalysts in the dataset (`valid' states) and $s'$ is reachable from $s$ via a valid action, $a$ (resulting in approximately 81,000 offline training tuples). The reward for a valid state is its inverse adsorption energy ($r = - E_{ads}$). We deem states with unknown adsorption energy `invalid' and assign them a penalty ($-\lambda$) reward. We use a DQN model \cite{mnih2015human} with 2 hidden layers of sizes 512 and 64 and hyperparameters similar to those used later in sections \ref{secRandomEdges} and \ref{secMulti} and train our network for 100,000 training steps.

We find that this approach does not yield a useful agent, and fails to noticeably converge towards desired states (Figure \ref{fig:exp1-2}). A high penalty is needed to coax the agent to stay away from invalid states: adsorption energies of valid states are in the range of $(0, -10)$ eV and we use a penalty $\lambda \in (10, 200)$. We find the agent mostly learns to move quickly to an invalid state and terminate with an overall reward of $-\lambda$, and avoids longer episodes that would accumulate multiple penalties. 

This can be understood by noting the sparsity of the state space. Referring to Figure \ref{tab:dataset}, we see that the cardinality of the `valid' set is around 2,379 catalysts, while it is $\simeq$160,000 for the whole space (all possible 3-element combos using one of 55 elements). This implies that over 98\% of possible states are invalid and we hypothesize that the sparsity of the data makes it too hard to learn to navigate to the optimal state. Figure \ref{fig:exp1-2} reports the low success rate for varying values of $\lambda$. 

\subsection{Simplified State \& Actions Space: Periodic Table of Elements}
\label{sec:exp2}
\paragraph{Experiment (2).}To test our hypothesis about the sparse reward issues faced in experiment (1), we aim to simplify our RL setup as much as possible to test whether agents trained in a more tractable setup do exhibit lower average terminal state energies. We implement the simplest environment for an agent to learn fundamental chemical knowledge: \emph{GridWorld} \cite{sutton2018reinforcement} on the Periodic Table of Elements \cite{mendeleev}. This setup is novel to our knowledge and can serve as a building block to learn more complex cheminformatics problems.

We define our Markov Decision Process (MDP) as follows:
\begin{itemize}
    \item 86 single element states (all elements from atomic number $Z=1$ (\ce{H}) to $Z=86$ (\ce{Rn});
    \item 5 actions: $\{ \_ | \leftarrow | \rightarrow | \downarrow | \uparrow \}$ to move between elements on the periodic table (see figure \ref{fig:periodicTableExp});
    \item Episodes last for a set 9 steps duration (long enough to reach the optimal element from any random starting point on the table).
\end{itemize}

We find that the Q-learning algorithm \cite{sutton2018reinforcement} on single elements with a reward defined as $E^2_{ads}$ consistently reaches the lowest energy states. Table \ref{tab:exp2} in Appendix reports final states for 20 random roll-outs: the agent reaches a -7.4 eV average terminal energy vs -1.5 eV for starting elements -- see figure (\ref{fig:periodicTableExp}) -- and terminates on top-2 states for $\sim~95\%$ of roll-outs starting from random states. We report 20 random roll-outs from a policy learned on this environment in Table \ref{tab:exp2} in appendix.

\subsection{Simplified Actions Space: Random Edge Traversal}
\label{secRandomEdges}

\paragraph{Experiment (3): Single Objective DQN with Random Edge Traversal on \emph{OC20}-Subgraph.}
In light of the results from experiment (2), we hypothesize that the periodic table environment works better than the original Section \ref{sec:exp1} setup for two reasons: (i) learning a limited set of actions is more tractable compared to the full 55 possible actions in experiment (1), and (ii) valid states are too sparse, and a negative reward on those invalid states leads to shorter episodes which never reach valid states. In other words, \textbf{we now traverse the subgraph of \emph{OC20}-subgraph materials with known adsorption energy for our target adsorbate}.

To that end, we reduce the original action space (from Section \ref{sec:exp1}) from ~60 actions to 5 actions, and introduce the following modified setup: \textbf{an action is only taken if it leads to a `valid' state}.  If it is invalid, the episode does not end and no reward is given; the state just remains unchanged.  Finally, we introduce Random Edge Traversal: instead of separately enumerating each element that could be added or removed, we add a random element if the agent chooses to expand the chemical compound. 

This method of constraining the action space is, to our knowledge, novel in the literature, and introduces a hybrid between bandits (pull a lever to add a random element) and reinforcement learning (the agent can backtrack, removing elements previously added). Formally, the Markov Decision Process (MDP) is defined as follows:
\begin{itemize}
\item A state is still a 55-dimensional 1-hot vector representing the catalyst chemical composition (up to three non-zero elements at any given time).
\item 5 possible actions: \texttt{<add>} a random element (that hasn't already been added to the catalyst), remove the \texttt{<first>}, \texttt{<second>}, or \texttt{<third>} element in the catalyst (if they exist), or \texttt{<stop>} the episode.
\item Rewards are returned when the agent terminates the episode or if we hit maximum episode length (between 20 and 75 steps). The reward function is $- E_{ads}^3$ to magnify the rewards towards the strongest (respectively weakest) binding elements.
\item Transitions are guided by the actions as in Section \ref{sec:exp1}, with the caveat that the state remains unchanged change when an invalid action is selected. Initial state is chosen at random, and discount factor is 0.9.
\end{itemize}

Our agent is a Double-DQN with two hidden layers with layer sizes 512 and 64. To avoid exploding Q-values explode when the target update period is too low, we use an update period of 300 steps, along with a learning rate of $10^{-3}$, and we implement an epsilon-greedy policy with $\epsilon=0.1$.

Experimental results reveal that trajectory roll-outs tend to be much longer than in Section \ref{sec:exp1}.  Common trajectory lengths range from 10-40 steps, while trajectories between 40 and 75 steps (the maximum length) are slightly less common. We report 9 random roll-outs from a policy learned on this environment in Table \ref{ref:exp3-1} in appendix. \textbf{We find that final states improve target adsorption energy compared to random initial states by an average of 4.1 eV.} 

We report average terminal state energies over 50 roll-outs for single-objective agents trained in experiments (1), (2) and (3) in table \ref{tab:exp1-2-3}. 

\begin{table}[!ht]
\begin{center}
\begin{tabular}{lcccc} 
Experiments  & Initial state & Exp (1) & Exp (2) & Exp (3) \\ \hline  \\
Avg. $E_{ads}$ (eV) & -1.5 & -2.2 & -7.4 & -5.6 \\ 
$\Delta$ (eV) & - & 0.7 & 5.9 & 4.1 \\ \\
\end{tabular}
\caption{Experiments (1), (2) \& (3) (single-objective). Average energies over 50 single-objective roll-outs. Objective: minimize adsorption energy (higher $\Delta$ is better).}
\label{tab:exp1-2-3}
\end{center}
\end{table}

\subsection{Multi-Element Multi-Objective DQN on \emph{OC20}-Subgraph}
\label{secMulti}

\paragraph{Experiment (4): Multi-Objective DQN.}

In practical catalysis experiments, a wide variety of potential adsorbates and possible reaction intermediates are present in the environment. Selectivity to a given reaction product is a major challenge \cite{norskov2009computational} \cite{de2019would}, and as a result, ideal catalyst design agents should learn to identify catalysts with stronger binding energy with certain adsorbates, and weaker adsorption with others. This is particularly important to try and break the `scaling relations' such as between $\star$\ce{O}, $\star$\ce{OH}, and $\star$\ce{OOH} in ORR \cite{man2011universality}.

We extend our experimental setup to a \textbf{multi-objective reinforcement learning problem}, where each adsorbate is a separate objective. In this setup, we train an agent to find catalysts that minimize (respectively maximizes) adsorption energies for each objective adsorbate. \textbf{We traverse the \emph{OC20}-subgraph of materials with known adsorption energy to at least one target adsorbate.}

We first implement a standard approach whenever we have multiple objectives: summing weighted rewards from each objective. In this scenario, the reward from each objective is the energy, and the weights are either -1 or +1, based on whether we seek to minimize or maximize the energy for that adsorbate. We train a DQN (same hyperparameters as previously), with the MDP defined as follows:

\begin{itemize}
\item A state is the same 55-dimensional 1-hot vector representing the catalyst compound.
\item 5 possible actions: \texttt{<add>} a random element, remove the \texttt{<first>}, \texttt{<second>}, or \texttt{<third>} element in the catalyst (if they exist), or \texttt{<stop>} the episode.
\item Rewards are returned when we either hit the maximum episode length, which is now set to 20.
The reward is the dot product of goal vector and adsorption energies (we don't use negative cube rewards since average adsorption energies matter for the maximization cases): 
$$r = - E_{ads} \cdot \mathbf{g}$$
\end{itemize}

\paragraph{Experiment (5): Multi-Objective DQN with Sub-Sampling.}
We hypothesize that as the number of objectives increases in the previous setup, there's a higher chance that the agent gets stuck in local minima, where a step in the environment may help optimize for one or two adsorbates but hurt the rest, which might discourage exploration. 

Considering that policies learned on a random mixture of objectives might encourage exploration, we introduce \textbf{Multi-Objective DQN with Sub-Sampling}, a new method by which we randomly sample one objective among the six for each training roll-out, :
$$r = -E_{ads_i} \times g_i \quad i_{\mathrm{roll-out}} \sim \mathcal{U}(i=1\dots6) $$

This roll-out-level objective sampling approach is, to our knowledge, a novel contribution to literature. We evaluate this approach in our experimental setup, by comparing sub-sampling pairs of objectives to a baseline from the previous experimental setup where all objectives are used in the reward computation for all roll-outs.

We report average final state energies over 50 roll-outs for experiments (4) and (5) in table \ref{tab:exp5} (stable over several runs). Final states found with sub-sampling are a better fit with their respective objectives for adsorbates 2, 3, and 4, but slightly worse for 1, 5, and 6. We notice that convergence with sub-sampled objective rewards takes longer than the baseline, as one may expect. Experimental results (reported in Table \ref{tab:exp5} and Figure (\ref{fig:multiElementMultiObj}) in appendix) \textbf{support the hypothesis that sub-sampling objectives helps encourage exploration, with longer average trajectories than baseline.} 

Overall, we find that, in the difficult multi-objective setting, \textbf{both baseline and sub-sampling approaches improve final state adsorption energy} towards the desired direction (increase or decrease), \text{simultaneously across all 6 adsorbates}, by an average of \textbf{0.8 eV}.

\begin{table}[!ht]
\begin{center}
\begin{tabular}{ l p{1.3cm} p{1.3cm} p{1.3cm} p{1.3cm} p{1.3cm} p{1.3cm} p{1.3cm} } 
Adsorbate  & 1: *CH2 & 2: *CH4 & 3: *N2 & 4: *NH3  & 5: *OH2 & 6: *OH \\ 
Objective & Increase & Increase & Increase & Decrease & Decrease & Decrease \\ \hline \\
Initial state & -2.2 & -3.3 & -1.8 & -1.6 & -1.9 & -1.9 \\ 
Exp (4): Baseline & \textbf{-2.2} & -3.0 & -1.5 & -1.9 & \textbf{-3.9} & \textbf{-3.9} \\ 
Exp (5): Sub-Sampling & -2.3 & \textbf{-3.0} &  \textbf{-1.6} & \textbf{-2.1} & -3.8 & -3.8 \\ \\
\end{tabular}
\caption{Experiment (4) and (5) (multi-objective). Average energies over 50 multi-objective roll-outs for the three experimental setups (rows) on each of the 6 objectives (columns).}
\label{tab:exp5}
\end{center}
\end{table}

\section{Analysis}

\subsection{Challenges of Sparse Known States Graph Traversal}
A particular challenging aspect of or our setup is the very sparse nature of the problem. The limited number of states for which adsorption energies are known ($\sim2,000$--$3000$) compared to the larger number of possible states ($55 + 55 \times 54 + 55\times54\times53 = 160,345$), makes learning to converge to low energy states difficult for our agent in the initial full state/actions setup. 

We found success after drastically simplifying our setup with the Periodic Table \emph{GridWorld} environment, which reinforced the intuition that limiting the number of actions is paramount to obtaining helpful results in a sprase rewards setup. This encouraged us to explore Random Edge Traversal of the \emph{OC20}-subgraph to significantly limit the number of actions a model must learn, and our results show this may prove a helpful general principle to traverse complex material spaces of sparsely documented properties.

\subsection{Generalizing Inverse Catalyst Design with RL}
Our results raise the question of whether the performance we report warrants the cost and complexity of training Deep RL models, where more standard optimization techniques would be more straightforward for offline learning on known energy datasets, even in the multi-objective setting.

Using Deep RL to solve this class of problems despite its inherent complexity enables us to address problems linear solvers cannot. While these experiments use offline learning on a dataset where adsorption energies are known, we envision using ML-based adsorption energy estimators as a critic in an actor-critic RL setup, and use exploitation of states with known energy to direct computational resources where exploring new unknown states is likely to be most beneficial.

Another enticing approach enabled by RL would be to include additional sparse signals to our reward model. For example, hard to model physical properties whenever they are known (stability, selectivity), materials cost, patents, or even human feedback (RLHF \cite{bai2022training}), where candidate catalysts are ranked by scientists based on their experience (e.g. manufacturing cost, experimental complexity, industry preferences, etc.). Such a complex reward function or multiple set of criteria would be hard to encode in a tractable way for a direct optimization setup to solve, but would be a good fit for multi-objective Deep RL approaches.

\section{Conclusion}

This paper presents our experiments with various Deep RL setups. We introduce Multi-Objective DQN with Sub-Sampling and Random Edge Traversal, a novel method to train a generalized multi-objective agent to traverse a large space of possible catalysts with sparse known properties, and identify materials with the desired adsorption energies profile for a set of target adsorbates of interest. 

We demonstrated that in the goal-conditioned multi-objective setting, Deep RL can identify promising materials for any combination of target adsorbates binding energies. In practice, conducting a large number of roll-outs and identifying the most common terminal states would point to promising materials on which to focus computational and experimental resources.

Our findings indicate the promise of Deep RL in navigating complex chemical spaces, and present novel approaches to tackle goal-conditioned multi-objective Reinforcement Learning for materials design. These methods could serve as a foundation for more complex computational challenges in large chemical spaces, and the development of novel materials for a wide range of applications in heterogeneous catalysis, electrochemical energy conversion, and low-emissions technologies.

\subsection*{Known Limitations and Future Work}

First, we report results on a single combination of 6 adsorbates, and a single objective vector in the  multi-objective setup. Further experimentations on a larger set of adsorbates objective vectors would help results robustness, especially for the novel objective sub-sampling approach we introduce. 

Second, we report goal-conditioning only on extrema, and train agents seeking to maximize or minimize adsorption energy. Other approaches to train our agent to seek states with intermediate adsorption energy, such as scalar conditioning (defining a target value for $E_{ads}$), would be particularly helpful to find optimal catalysts, which usually have intermediate binding strength on the activity-binding energy `volcano plots' (Sabatier Principle \cite{norskov2009computational}). 

Relabeling targets is a potent way to improve goal-conditioned agents, and future work could focus on improving the performance of our multi-objective agent through Hindsight Experience Replay \cite{andrychowicz2017hindsight}, as well as Prioritized Experience Replay (PER) \cite{schaul2015prioritized}. Other offline algorithms such as Conservative Q-Learning \cite{kumar2020conservative} may also prove helpful.

Lastly, our overall objective was to train an RL agent to traverse a vast space of possible materials with multiple target adsorbates. While we used offline RL on known adsorption energies datasets to facilitate experimentation, we envision an actor-critic setup where an actor is trained using our graph traversal setup, and a critic uses ML-based adsorption energy evaluation models. An ML-based, lightweight critic such as \emph{AdsorbML} \cite{AdsorbML} could perform approximate but fast energy evaluations for unknown states traversed at roll-out time, and more exact DFT computations could be reserved for frequent final states revealed by accumulated roll-outs as strong catalyst candidates.

\subsection*{Code \& Data Access}
The code implementation for our experiments, as well our datasets compiled from the \textit{Materials Project} \cite{MaterialsProject} and \textit{Open Catalyst 2020} \cite{OCP} datasets, are made available for download for reproducibility purposes on the \href{https://github.com/rlacombe/AdsorbRL/tree/main/data}{AdsorbRL GitHub repository}.

\subsection*{Acknowledgements}
The authors wish to thank Prof. Chelsea Finn, Dr. Karol Hausman, and Jonathan Yang at Stanford University for guidance on this project, as well as Ajay Kannan for his contribution to Random Edge Traversal and the multi-element, multi-objective setup. We are grateful to Meta AI and the chemical engineers, materials scientists and AI researchers who collaborated on the \textit{Open Catalyst Project} and \textit{Materials Project}, and everyone whose research informed these data sets. This work would not have been possible without them.

\appendix

\bibliographystyle{unsrt}
\bibliography{bibliography}

\clearpage

\begin{table}[!ht]
\begin{center}
\begin{tabular}{  l c  } 
 Adsorbate & \begin{tabular}[c]{@{}l@{}}Dataset size \\ (\# of known catalysts)\end{tabular}  \\ \\ \hline \\
 $\star\ce{OH2}$ & 2,379  \\
 $\star\ce{CH2}$ & 2,759  \\
 $\star\ce{CH4}$ & 2,409  \\
 $\star\ce{N2}$ & 2,111   \\
 $\star\ce{NH3}$ & 2,473   \\
 $\star\ce{OH}$ & 2,655  \\
All adsorbates & 7,386 (unique) \\  \\
\end{tabular}
\caption{Sparse rewards: number of catalysts for which adsorption energy is known for the corresponding adsorbates, out of $\sim$160,000 possible compounds. Does not sum up due to duplicates.}
\label{tab:dataset}
\end{center}
\end{table}

\begin{figure}[ht]
    \centering
    \includegraphics[width=\linewidth]{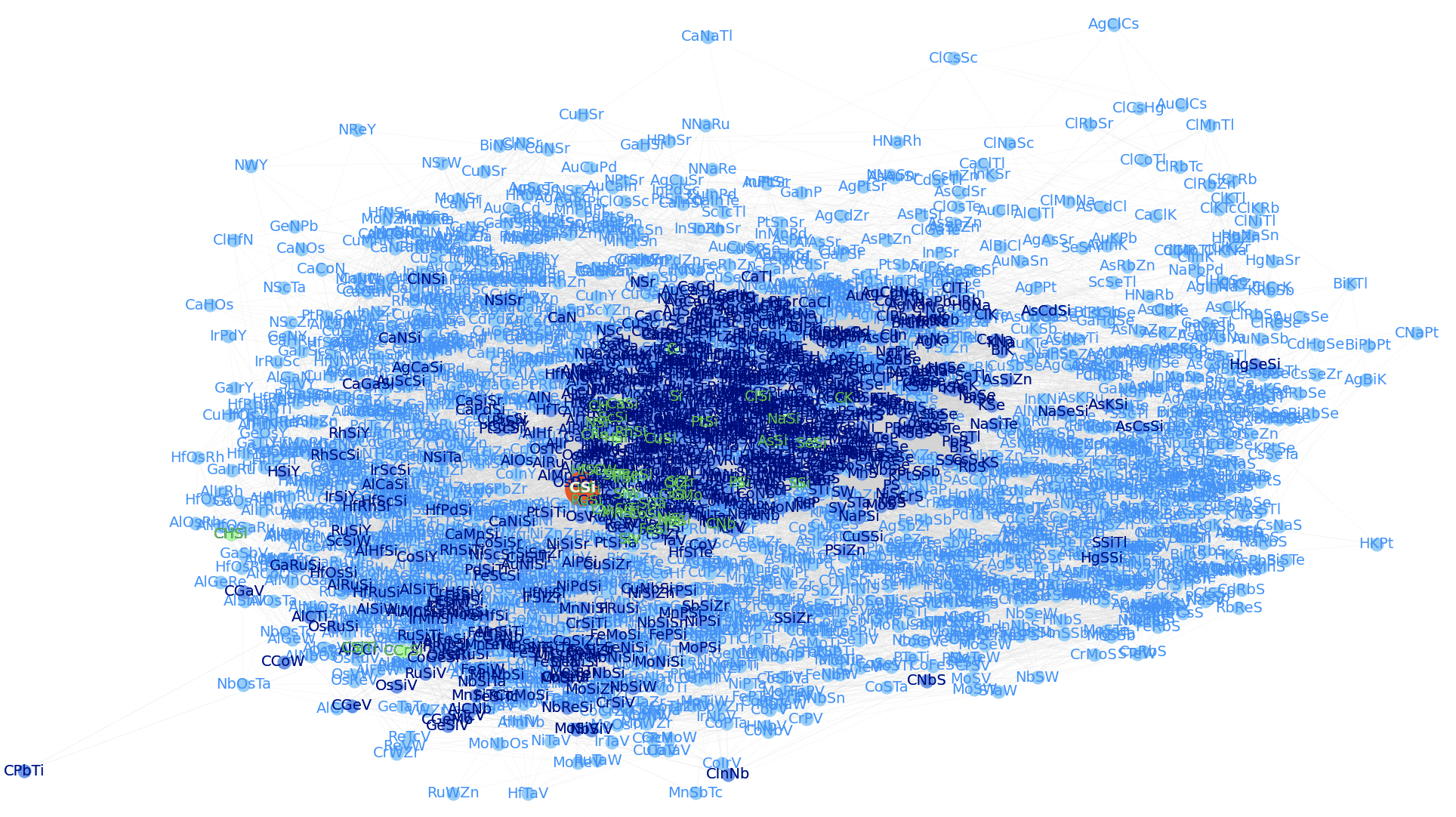}
    \caption{3-hop ego graph of the lowest energy state for $\ce{*OH_2}$ adsorbate (SiC). Red: SiC. Green: 1-hop ego network. Dark blue: 2-hope ego network. Light blue: 3-hop ego network.}
    \label{fig:elementGraph}
\end{figure}


\begin{figure}[ht]
\centering
    \includegraphics[width=0.9\linewidth]{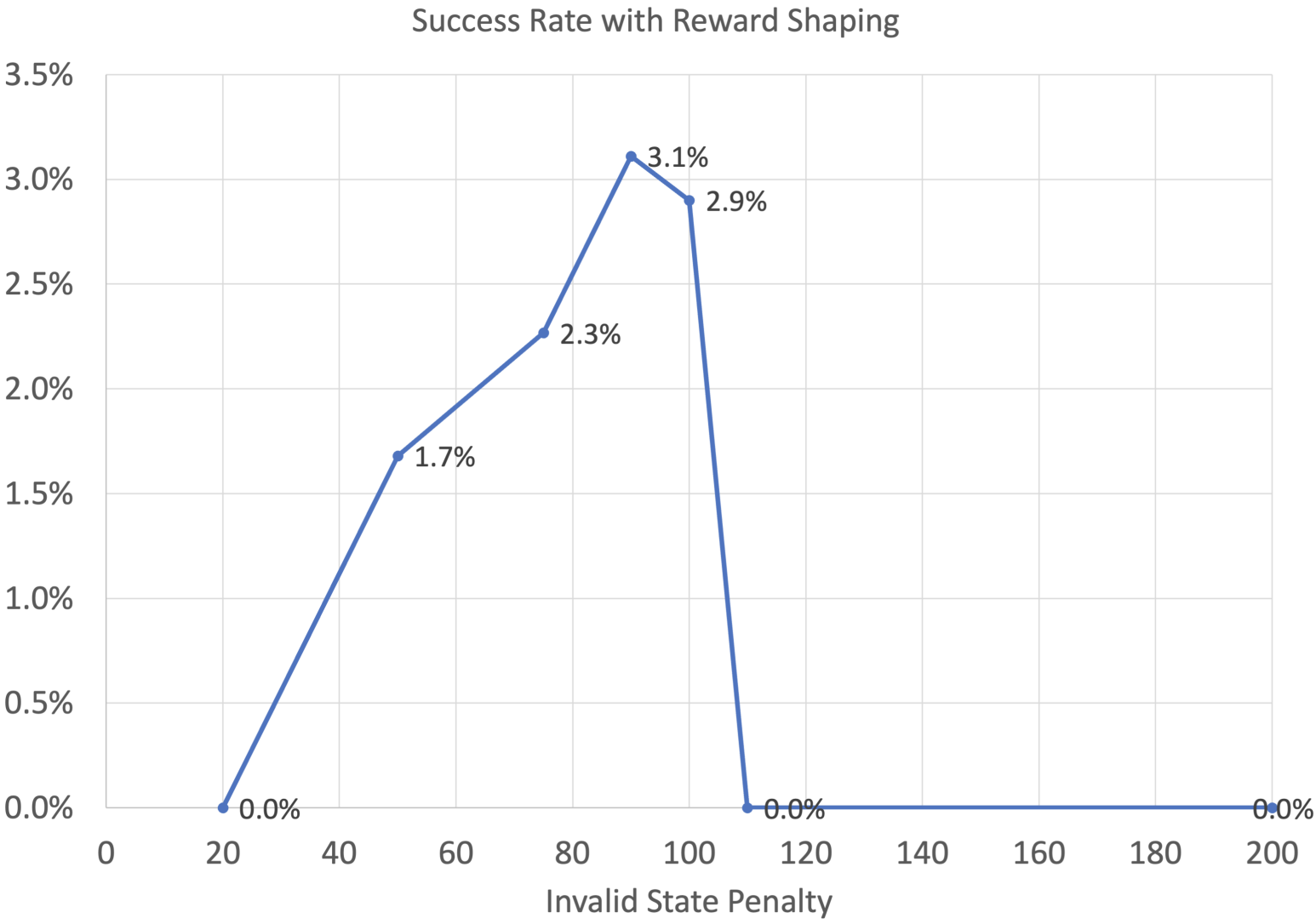}
\caption{\textbf{Experiment (1).} Success rate in reaching the optimal valid state using reward shaping for each value of $\lambda$. The graph illustrates the difficulty in training a DQN agent that converges in our dataset with sparse rewards.}
\label{fig:exp1-2}
\end{figure}

\begin{figure}[!ht]
\centering
    \includegraphics[width=1\linewidth]{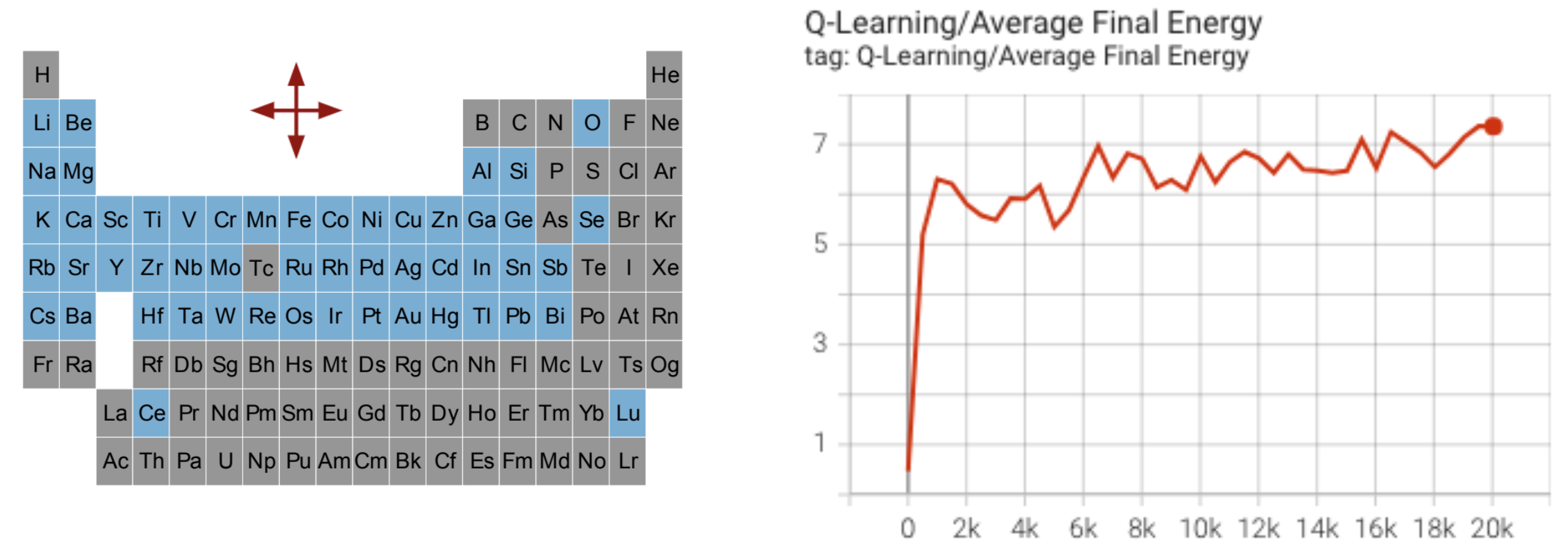}
\caption{\textbf{Experiment (2).} Left: Periodic table of elements, traversed in a \textit{GridWorld} setting (stay | left | right | up | down). Elements present in \emph{OC20} are highlighted in blue. Right: average final energy after evaluation roll-outs while Q-learning the policy (NB: plot represents $-E_{ads}$, final adsorption energy is negative).}

\label{fig:periodicTableExp}
\end{figure}

\begin{table}[!ht]
\begin{center}
\begin{tabular}{  c c c  } 
 Final state & Trajectory length & Energy (eV) \\ \hline
 Carbon & 9 & -7.9 \\ 
 Carbon & 9 & -7.9 \\ 
 Iron & 9 & -9.1 \\ 
 Carbon & 9 & -7.9 \\ 
 Iron & 9 & -9.1 \\ 
 Carbon & 9 & -7.9 \\ 
 Iron & 9 & -9.1 \\ 
 Manganese & 9 & -0.9 \\
 Carbon & 9 & -7.9 \\ 
 Carbon & 9 & -7.9 \\ 
 Carbon & 9 & -7.9 \\ 
 Carbon & 9 & -7.9 \\ 
 Carbon & 9 & -7.9 \\ 
 Carbon & 9 & -7.9 \\ 
 Carbon & 9 & -7.9 \\ 
 Iron & 9 & -9.1 \\ 
 Iron & 9 & -9.1 \\ 
 Carbon & 9 & -7.9 \\ 
 Carbon & 9 & -7.9 \\  \\
 
\end{tabular}
\label{tab:exp2}
\caption{\textbf{Experiment (2).} Twenty random roll-outs from a policy learned on this environment.  Lower energy is better.  Average random single-element reward is -1.5eV.}
\end{center}
\end{table}

\begin{table}[!ht]
\begin{center}
\begin{tabular}{  c c c  } 
 Final state & Trajectory length & Energy (eV) \\ \hline
 Carbon & 24 & -7.9 \\ 
 Iron & 40 & -9.1 \\ 
  Sulfur and Vanadium & 5 & -6.3 \\
 Iron & 10 & -9.1 \\ 
  Iron & 24 & -9.1 \\ 
 Cesium and Hydrogen & 1 & -0.8 \\ 
  Tantalum and Vanadium & 9 & -2.1 \\ 
 Iron & 24 & -9.1 \\
  Iron & 18 & -9.1 \\  \\
\end{tabular}
\label{ref:exp3-1}
\caption{\textbf{Experiment (3).} Nine random roll-outs from a policy learned on this environment.  Lower energy is better.  Average random single-element reward is -1.5eV.}
\end{center}
\end{table}

\begin{figure}[!ht]
    \centering
        \includegraphics[width=0.8\linewidth]{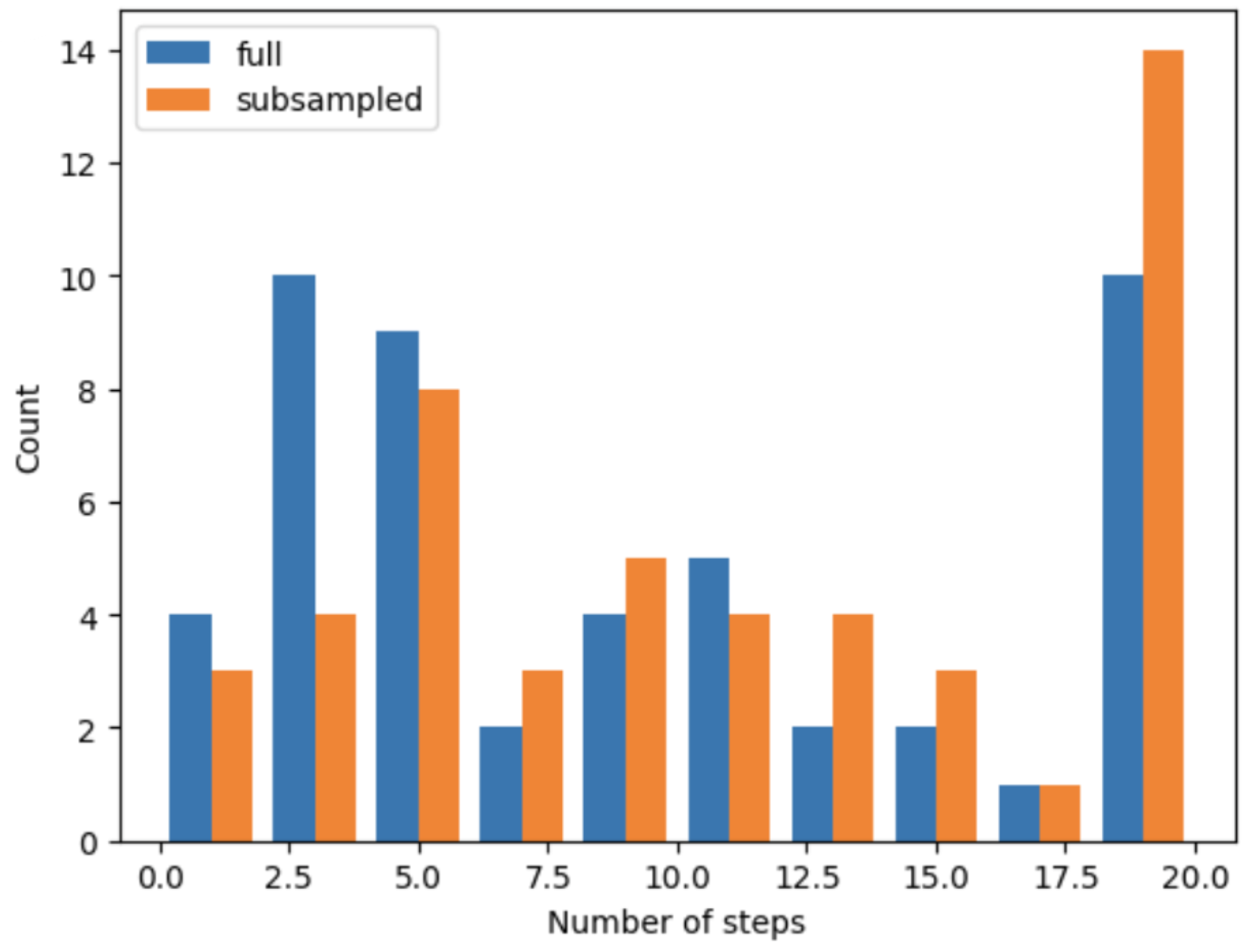}
    \caption{\textbf{Experiments (4) \& (5): multi-objective DQN on OC20-Subgraph.} Number of episode steps at roll-out, full objective vs objective sub-sampling.}
    \label{fig:multiElementMultiObj}
\end{figure}

\end{document}